\documentclass{llncs}
\usepackage{graphicx}
\usepackage{amsmath,amssymb} 
\usepackage{color}
\begin{document}

\mainmatter

\title{Accurate and Robust Neural Networks for Security Related Applications Exampled by Face Morphing Attacks}

\authorrunning{C. Seibold\inst{1}, W. Samek\inst{1}, A. Hilsmann\inst{1} and P. Eisert\inst{1,2}}
\titlerunning{Accurate and Robust Neural Networks for Security Related Applications}

\author{Clemens Seibold\inst{1}, Wojciech Samek\inst{1}, Anna Hilsmann\inst{1} and Peter Eisert\inst{1,2}}

\institute{Fraunhofer HHI, Einsteinufer 37, 10587 Berlin, Germany
\and Humboldt University Berlin, Unter den Linden 6, 10099 Berlin, Germany}

\maketitle

\begin{abstract}
Artificial neural networks tend to learn only what they need for a task. A manipulation of the training data can counter this phenomenon. In this paper, we study the effect of different alterations of the training data, which limit the amount and position of information that is available for the decision making. We analyze the accuracy and robustness against semantic and black box attacks on the networks that were trained on different training data modifications for the particular example of morphing attacks. A morphing attack is an attack on a biometric facial recognition system where the system is fooled to match two different individuals with the same synthetic face image. Such a synthetic image can be created by aligning and blending images of the two individuals that should be matched with this image.

\end{abstract}

\section{Introduction}
Biometric verification systems are nowadays present in many fields of daily life. They are used in the consumer market, e.g. to unlock a mobile phone, as well as for sovereign tasks like automatic border control. One huge advantage and a reason for their success is that every individual carries his or her biometric characteristics always with them and that they are difficult to copy by another individual. 
Recently, the vulnerability of facial recognition systems against a specific attacking method was made known \cite{Ferrara14}. This attack, which is called morphing attack, tricks a facial recognition system into matching two different individuals with one forged facial image. The detection of this kind of fraud is substantial to ensure the trustworthiness, especially for security-related tasks like entrance control.

Recently, several attacks on the prediction of neural networks with only small perturbations of the content and without knowledge of the weights or architecture of the network were published. This represents a danger for neural networks in many applications, especially for security related applications where the network should be accurate and robust against specific attacks on their decision making process. By the example of morphing attack detection, we present in this paper different ways to evaluate the properties of neural networks, e.g. robustness against attacks, and the effect of different modification of the training data on these properties. We train VGG19 \cite{VGG19} networks to detect morphing attacks and differ the amount and location of information that can be used by the network to detect the forgeries. The differently trained networks are analyzed regarding their accuracy and two different kinds of attacks on the decision making. One attack scenario represents a semantic attack where the forger leaves only a few traces in the image so that the network needs to be able to recognize every kind of artifacts to detect a morph. In the other scenario, we assume that the attacker has access to the system as a black box and can test the system as often as he likes. In order to get insights on the decision making of the different networks and to understand the effects of the training data alteration, we use layer-wise relevance propagation \cite{BachPLOS15}.

In the next section, we give a summary of the recent studies on face morphing attacks and attacks on neural networks. Our fully automatic morphing pipeline is presented in section 3, followed by details about our self-made and collected face and morphing attack database in section 4. The setup of our experiments and thus the generation of our specifically altered training data is described in section 5. The analysis methods and results are presented in sections 6 and 7.

\section{Related Work}
The vulnerability of facial recognition systems was first published by Ferrara et al. \cite{Ferrara14}. They created morphing attacks by manually aligning facial images of different individuals and showed that they can trick commercial facial recognition systems to verify both individual with the forged face image.

Several approaches for morphing attack detection were developed based on different techniques. 
Raghavendra et al. \cite{Raghavendra16} extracted facial micro-textures using Binarized Statistical Image Features to distinguish between morphing attacks and genuine images. The morphing detection method of Makrushin et al. \cite{MakrushinND17} is based on JPEG double compression. Neubert studies in \cite{Neubert17} image degeneration to detect morphing attacks. Morphing detectors based on CNNs were successfully used by Seibold et al. \cite{SeiboldSHE17} and Raghavendra et al. \cite{Raghavendra17}.
Ferrara et al. \cite{Ferrara18} went even further and proposed a demorphing method, which is capable of extracting one genuine face of a morphing attack, if an image the other face is known.

The vulnerability of neural networks against small perturbations in the image was revealed by Szegedy et al. \cite{Szegedy14}. They also showed that the required perturbations are not of random nature and depend on the dataset. To get these perturbation they used a gradient-based method and backpropagation.
Goodfellow et al. \cite{Goodfellow15} proposed the fast gradient sign attack that requires only the sign of the gradients of the required perturbation to perform the attack.
Papernot et al. \cite{Papernot17} published a method to estimate the signs of these gradients without having access to the weights or architecture by training a substitute model that behaves similar to the network that should be fooled. To do so, they needed only access to the final output of the network for generated images. Thus, they provided a framework to use the fast gradient sign attack as a black box attack. Other black box attacks that also try to model the network's behavior by questioning the network were published by Narodytska and Kasiviswanathan \cite{NarodytskaK17} and Dang et al. \cite{Dang17}. Su et al. \cite{Su17} showed that even a larger modification of only one pixel is sufficient to change the prediction of the network.

\section{Forged Facial Image Generation}

\subsection{Overview}
\begin{figure}
\center
\includegraphics[width=12cm]{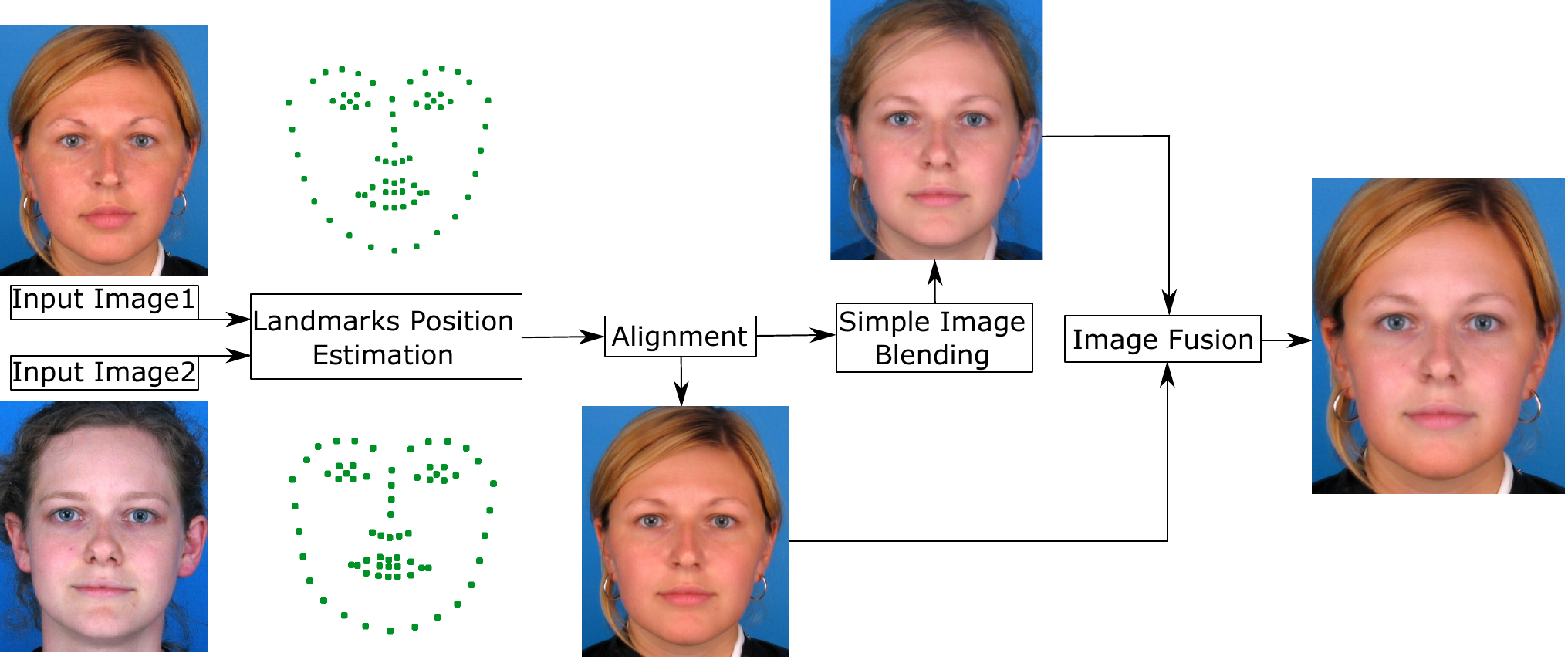}
\caption{Morphing Pipeline.}
\label{fig:pipeline}
\end{figure}

Since we need a lot of morphing attacks to train our networks, we implemented a fully automatic morphing pipeline similar to the face morphing pipelines introduced in \cite{MakrushinND17,SeiboldSHE17}. The idea of all of these pipelines is to create a facial image that contains characteristics of two different individuals. They all start with an alignment of two facial images and use additive alpha blending to generate the forgery. Their differences are mostly in the processing of regions that are not part of the inner face. Figure \ref{fig:pipeline} provides an overview about our face morphing pipeline. Our alignment and blending is presented in the following.

\subsection{Image Alignment}
We align the facial images based on predefined facial landmarks whose positions we estimate using dLib's \cite{dlib} implementation of the facial feature detector described in \cite{Kazemi14}. The images are warped such that the facial landmarks of both images are at the same position and the regions between landmarks are transformed similar to the nearest facial landmarks. The target position of a landmark is its average position of both input images.

In the following, we describe two different warping methods, whereby we refer to the original image as source image and to its warped version as target image and accordingly to the landmark positions as source positions and target positions. Both approaches use different numbers of facial landmarks. We add some additional landmarks for the triangle warp, whereas the field warping approach only uses a small subset of the estimated landmarks. We used two different kinds of warping methods to increase the variety in data generation and reduce a possible impact of the warping algorithm on our experiments.
\begin{figure}[htbp]
\begin{center}
\includegraphics[width=3.5cm]{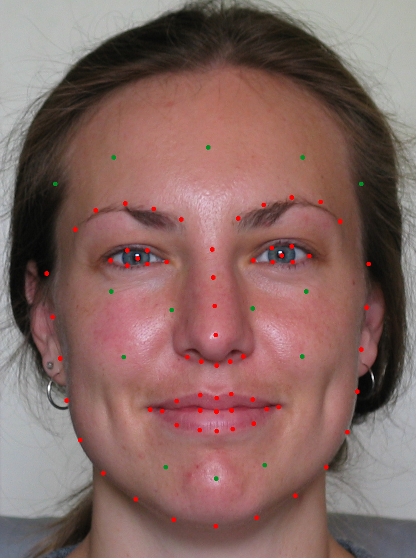} \hspace{0.4cm}
\includegraphics[width=3.5cm]{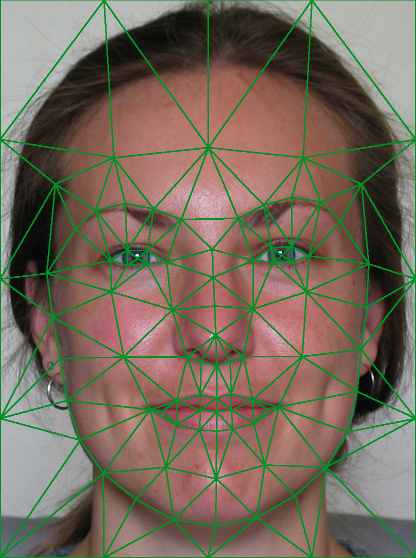} \hspace{0.4cm}
\includegraphics[width=3.5cm]{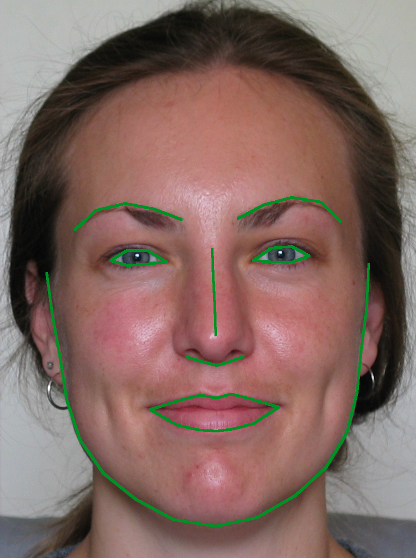}
\end{center}
\caption{Basis of our image warping methods: landmarks estimated by dlib in red and additional added landmarks in green (left), triangle mesh for triangle warp (center), lines for field morphing (right).}
\label{fig:feats}
\end{figure}
\paragraph{Triangle Warp}
We add and fuse some landmarks to get a triangle mesh that includes the whole image and to avoid degenerated triangles. The additional landmarks are at the borders of the images, in the center of each eye, on the cheeks, between mouth and chin and on the forehead, as shown in figure \ref{fig:feats}. The points on the upper and lower line in the center of the lips, which were estimated by dLib, are fused, since they are very close. After adding and removing the feature points, we apply the Delaunay triangulation on the target point set to obtain a mesh topology. This topology is then applied on the source point set. Now, having a triangle mesh with the same topology on each image, we warp the region of each triangle in the source image to the region of the triangle in the target image using an affine transformation.

\paragraph{Beier-Neely field morphing}
The Beier-Neely field morphing algorithm \cite{Beier92} requires corresponding line segments between the images to define a transformation. Since we only estimate a set of points, we connect some of them with a fixed predefined pattern, as shown in figure \ref{fig:feats}, to be able to apply the field morphing algorithm. The concept of this morphing method is to transform each pixel relative to the transformation of each line segment. The closer a line segment is to a pixel, the stronger is its impact on the pixel's transformation. The pixel position of a target pixel in the source image is calculated by applying the transformation for each line pair individually, adding the weighted displacement vector and normalizing it by the sum of weights. For one line segment $Q$ in the target image with end points $Q_s,Q_e$ with $Q_n$ being the normal of the line segment, a pixel $\mathbf{p}$ can be parametrized with two parameters $u,v$ by
\begin{equation}
\mathbf{p} = u \|Q_e - Q_s\| + Q_s + Q_n v.
\end{equation}
The pixel position $\mathbf{p}'$ in the source image can now be calculated by
\begin{equation}
\mathbf{p'} = u \|Q'_e - Q'_s\| + Q'_s + Q'_n v,
\end{equation}
with $Q'_e,Q'_s, Q'_n$ being the end points and normal of the corresponding line segment in the source image.
The impact or weight of a line on a pixel is defined with respect to their distance in the target image by
\begin{equation}
w(\mathbf{p}, Q) = \left(\frac{1}{d + \varepsilon}\right)^2,
\end{equation}
with $d$ being the shortest distance from the pixel $\mathbf{p}$ to the line $Q$.

\subsection{Image Blending}
Since most of our facial landmarks are located in the face and not around the head, ears or hair, the  alignment procedure takes only the inner part of the face below the forehead into account. A simple blending of the whole image would thus lead to strong visible ghosting artifacts around the silhouette, hairstyle, ears and, if included, shoulders. To avoid this  striking forgery indicators, we used the blending method proposed by Seibold et al.\cite{SeiboldSHE17}.
We blend only the inner part of the face and use one aligned input image for the outer part. As a consequence, we have to calculate a smooth and inconspicuous transition from the blended image to the warped original one. The area of this transition is defined as region between two ellipses with the same center. Their width and height are determined relative to the estimated facial landmark positions. As the authors of \cite{SeiboldSHE17} do, we also separate the images in a high spatial frequency and a low spatial frequency image and calculate for both of them a different transition between the inner and the outer part. In this way, we can avoid cuts through high frequency differences in the images and low frequency differences caused by slightly different skin colors or different illuminations can be handled by a smooth and small color gradient over this zone. The transition for the low spatial frequency part is calculated using Poisson Image Editing \cite{Poisson03}. The gradients for the Poission equation in the transition zone are taken from the blended image. For the high spatial frequency part of the images, we calculate a cut through the transition area that does not create high frequencies around the cut. For that, we use the graph cut algorithm with the penalty term used in \cite{SeiboldSHE17} that prefer cuts through regions with similar gradients.
As the authors of \cite{SeiboldSHE17}, we apply this algorithm on images in polar coordinates, so that shorter parts around the inner border do not have lower cost due to less cuts.
\section{Training Dataset}
\subsection{Database}
To get a representative database, we acquired several freely-available face databases and captured face images by ourselves. All of these images were manually checked whether they are suitable for passports and images were removed in which the person looked not directly into the camera, had the eyes closed, the mouth opened or other requirements for passports were strongly violated. We gathered about 1,900 images of different individuals. Table \ref{tab:faceImageSources} provides an overview about the image sources and number of used images per database.\\
\begin{table}[!htpb]
\center
\caption{Face image sources.}
\begin{tabular}{cccccccccc}
Database & \ BU-4DFE\cite{BU4DFE}\ &\ CFD\cite{CFD}\ &\ FEI\cite{FEI}\ &\ FERET\cite{FERET}\ &\ PUT\cite{PUT}\ \\
\hline
\#used images &\ 101\ &\ 597\ &\ 100\ &\ 459\ &\ 100\ 
\end{tabular}\\[1.5ex]
\begin{tabular}{cccccccccc}
Database & \ scFace\cite{scFace}\ &\ Utrecht\cite{utrecht}\ &\ own &\ various\ \\
\hline
\#used images&\ 108\ &\ 64\ &\ 261\ &\ 89\ 
\end{tabular}
\label{tab:faceImageSources}
\end{table}
We divide our set of original images into a training set that contains 80\% of all original images, a testing set (15\%) and a validation set (5\%). This distribution into training, testing and validation sets hold approximately also for each individual database. 
Since there is obviously a strong dependency in the data between original images and morphs, we used for the generation of morphing attacks for the training, testing or validation set only images from the same set. 
To ensure a good quality and variety of the morphs, we made some guidelines for the selection of pairs of original images for morphing attacks:
\begin{enumerate}
	\item both individuals have the same gender
	\item 50\% of the morphs use the triangle approach and 50\% the field morphing
	\item both images are from the same data base
	\item each image is approximately used equally frequent to create a morph
\end{enumerate}

\subsection{Preprocessing}
The generation of a morph requires a lot of image manipulations that change characteristics like gradient distribution, correlation between neighbored pixels and so on. This creates a lot of low-level artifacts that might be used by the network to discriminate a morph from an original image \cite{SeiboldSHE17}. Since this characteristics depend strongly on the input data, we apply some kinds of noise and blur to change this distributions in the original and forged face images to prevent the network from discriminating on these characteristics and to focus on more semantic high-level models. For each original image and for each morphing attack in a set, we added two different kinds of noise and applied two different kinds of blurring filters. Thereby, each image is added in five different versions to one set: one version without any processing and one each after applying motion blur, Gaussian blur, Salt-and-pepper noise or Gaussian noise. The parameters for the different kinds of blur and noise vary for each image  as shown in table \ref{tab:ProProc}.
\begin{table}[!tpb]
\caption{Parameters for data preprocessing.}
\begin{tabular}{l | l}
Kind of Preprocessing &\ Parameter Range\\
\hline
motion blur &\ length between 0.5\% and 1.0\% of image height, random angle \\
Gaussian blur &\ sigma between 0.25\% and 0.5\% of image height\\
Salt-and-pepper noise &\ 1\% of all pixels\\
Gaussian noise &\ standard deviation of 0.05 (pixel intensities in [0,1])
\end{tabular}
\label{tab:ProProc}
\end{table}
To remove unnecessary variance in the input data and have a standardized input for the network, we rotate, rescale and crop the images before feeding them into the network. Each image is rotated such that the eyes are at the same height and finally scaled and cropped to a size of $224\times224$ such that it contains the region between eye brows and mouth and between the outer pars of the eyes. During the training, this region is shifted slightly to augment the training data.

\section{Experiment Settings}
\begin{figure}
\center
\includegraphics[width=0.2\textwidth]{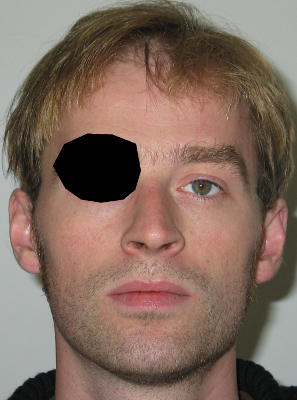}
\includegraphics[width=0.2\textwidth]{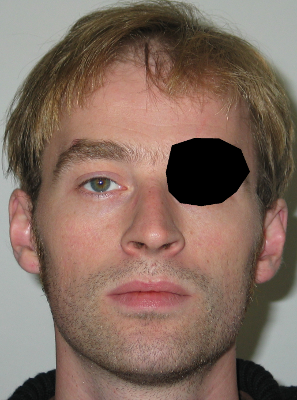}
\includegraphics[width=0.2\textwidth]{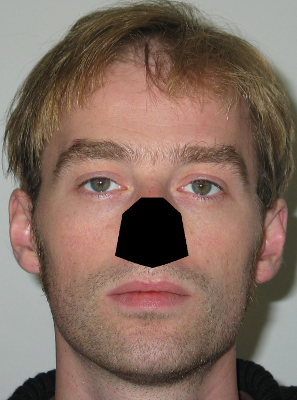}
\includegraphics[width=0.2\textwidth]{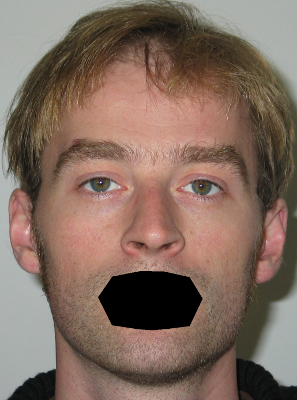}
\caption{Regions for partial morphs.}
\label{fig:partialMorphRegions}
\end{figure}

We start our training in all cases with a VGG19 network \cite{VGG19} that was pretrained on the ILSVRC challenge. Although a naive training with genuine images and morphing attacks leads to a quite good accuracy, its robustness against specific attacks is rather poor, as shown in the following sections. In order to train a network that is accurate and robust, we use different alterations of our training data. Our key idea is to change the amount, location and kind of information that is available to detect a morph. In this way we force the network to consider all regions of the face during the training. For this purpose, we use partial morphs. A partial morph is a forged facial image in which some regions are taken from a morphed image and the rest from an aligned input image. We defined four different regions in the face for the generation of our partial morphs. These regions are the region around the right or left eye containing the eyebrows, around the nose and around the mouth, see also figure \ref{fig:partialMorphRegions}. For each morphing attack, the regions were automatically estimated relative to the facial feature points estimated by dLib.

In the following, we present the four different kinds of training data manipulation that were used in our experiments. In three cases, we perform a two class classifier training. If an image has at least one forged region, the whole image is assigned to the class morphing attack. In the fourth case, we first do a multiclass training with four different classes. Each class represents one region and if this region is forged, the image belongs to this class. To get a binary decision, we remove the last fully connected layer and train the last layer and with a smaller learning rate the second last layer for the binary class problem using the complex morphs training method described below.\\
The relative amount of morphs and genuine image is designed such that each class occurrences with equal frequency. The distribution of the different types of morphs and the amount of genuine images for the different training methods is as follows:\\
\textit{Naive training: }
50\% genuine images, 50\% complete morphs\\[0.75ex]
\textit{One region: }
50\% genuine images, 10\% complete morphs and 4 $\times$ 10\% with one region morphed\\[0.75ex]
\textit{Complex morphs: }
50\% genuine images, 10\% complete morphs, partial morphs with 10\% one, two, three and four region morphed (each region is morphed with equal frequency)\\[0.75ex]
\textit{Multiclass (Complex morphs): }
 partial morphs with zero, one, two, three or four morphed regions (each 20\% and each region is morphed with equal frequency); A two class classifier is obtained by reinitializing the last fully connected layer and training it with the same data as used in the complex morphs training method.
  
As mentioned above, the aim of our training data modification is to force the network to take all available information for its decision making into account. In case of a naive training it might be sufficient to focus on the eyes to detect morphing attacks, but there are still other relevant information available that were ignored by the network. In case of partial morphs the network has to gather information from all region in the face to make a correct class prediction. One disadvantage of the one region morphed training method is that less artifacts are presented to the network during the training, if we keep the amount of genuine images and forgeries identical. The complex morphs training method overcomes this disadvantage partly, but again the network does not have to gather all available information in one morph to classify it correctly. The multiclass complex morphs training method allows to present more artifacts and keeps the number of images that are in one class identical to the images that are not in this class. This way, we can present more morphing artifacts to the network without creating a class bias and can still force the network to use all available information about morphing artifacts in one image.
\section{Analysis Methods}
We analyze the differently trained networks based on the accuracy and different measurement for robustness. Especially a network for forgery detection should be robust against all kind of attacks on its decision. Thus, we study two different problem specific attacks, in which the attacker knows nothing about the forgery detection system but tries to manipulate the images as little as possible and as good as possible to leave no suspicious traces, and general attacks on neural networks, where the attacker has access to the system as a black box.\\
Under the assumptions, that a professional tries to avoid artifacts and leaves only some of them, the detection of partial morphs seem to be a good criteria for robustness.\\
In another attack scenario, the attacker has access to the system and can test the forged images as often as possible.  Thus, he can attack the decision of the network directly by creating adversarial examples that differ only slightly to the created morphing attack but are classified as genuine image by the neural network. \\
We modeled this attack using the fast gradient sign method \cite{Goodfellow15} and generated the adversarial examples using a substitute model \cite{Papernot17}. This model is trained on our validation data set and on adversarial examples that were generated as in \cite{Papernot17}. Since we had no success in fooling our networks using the shallow networks in \cite{Papernot17}, we use a pretrained GoogLeNet \cite{GoogLeNet} instead.\\
Beside our accuracy and robustness analysis, we try to get some insights on the networks decision making and whether or how they differ \cite{SamITU18b}. For thus purpose, we use the LRP-Toolbox \cite{LRP} to mark regions in the image that were relevant for the network's decision. Over-simplified described, a region is relevant, if at least one path from a neuron in the last layer that defines the class affiliation to this region exists, and the neurons along this path have a strong activation and large weights.
We apply the LRP-toolbox on partial morphs to check if the relevance is mostly in the morphed region and thus the classification was done by detecting morphing artifacts. We use the epsilon decomposition as relevance propagation method for the fully connected layers. The relevance of the later convolutional layers was propagated using the $\alpha-\beta-\text{decomposition}$ with an $\alpha$ of $2$ and $\beta$ of $-1$ and the relevance of the first convolutional layers using flat weight decomposition. We consider only forgeries that were classified as such or with an output at the softmax for the class "morphing attack" was of least 0.1 so that there is relevance to propagate to the image.
\section{Results}
In the following, we present the accuracy and robustness that is achieved with the different kinds of training methods and discuss the insight we got using relevance propagation.\\
\subsubsection{Accuracy}
We refer to the relative amount of correctly classified genuine images as true positive rate and to the relative amount of correctly classified morphing attacks as true negative. The best true positive rate was achieved using the naive training method. This might be due to the fact that genuine images and morphing attacks share the same content for partial morphs which were used in the other training methods. All networks were better in classifying morphing attacks than genuine images. By moving the threshold for the classifier, we could counter this bias and got for the naive and the multiclass complex training method the best equal error rates, whereas the EERs for the other two training methods were at least twice that much.
\begin{table}[!htpb]
\center
\caption{Accuracy of different training methods.}
\begin{tabular}{l|cccc}
&\ naive\ &\ one morphed\ &\ complex morphs\ &\ multiclass\\
\hline
true positive & 95\% & 90\% & 93\% & 92\%\\
true negative & 98\% & 95\% & 95\% & 99\%\\
EER & 3.1\% & 7.2\% & 6.1\% & 2.8\%
\end{tabular}
\end{table}
\vspace*{-0.5cm}
\subsubsection{Semantic Attacks}
Table \ref{tab:RobustnessPartialMorphs} shows the true negative rates of our differently trained networks for different kinds of partial morphs, where the attacker blends only the region around right or left eye, the nose or the mouth. Our naive trained network performs quite poor for this scenario, whereas all other training methods perform significantly better. The detection rate for the same network but for different kinds of partial morphs differ significant for some networks. The naive training method performs worse in relative terms for partial morphs of the nose and mouth, whereas the other three methods perform only worse on the mouth than on partial morphs of other regions.
\begin{table}[!htpb]
\center
\caption{Robustness against partial morphs.}
\begin{tabular}{l|ccccc}
&\ left eye\ &\ right eye\ &\ nose\ &\ mouth\ &\ average\\
\hline
naive 			& 25\% & 21\% & 14\% & 13\% & 20\% \\
one morphed		& 81\% & 89\% & 79\% & 71\% & 80\%\\
complex morphs  & 78\% & 74\% & 73\% & 54\% & 70\%\\
multiclass 	    & 86\% & 93\% & 90\% & 79\% & 87\%
\end{tabular}
\label{tab:RobustnessPartialMorphs}
\end{table}
\vspace*{-0.5cm}
\subsubsection{Black Box Attacks}
\begin{figure}[!htpb]
\center
\includegraphics[width=0.23\textwidth]{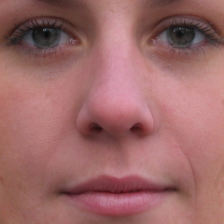}
\includegraphics[width=0.23\textwidth]{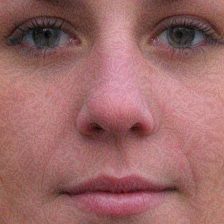}
\hspace{0.1cm}
\includegraphics[width=0.23\textwidth]{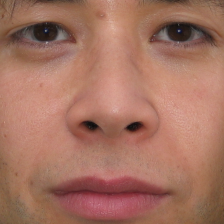}
\includegraphics[width=0.23\textwidth]{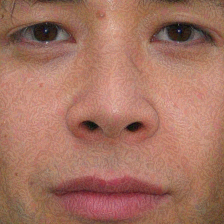}
\caption{Adversarial examples created for the naive trained network. The left image is in both cases the original image and the right image the adversarial attack with a change of the intensities of 6. The attack on the right was successfull, whereas the adversarial attack on the left was still classified as fraud.}
\label{fig:DNNAttacksNaive}
\end{figure}

\begin{figure}[!htpb]
\center
\includegraphics[width=0.9\textwidth]{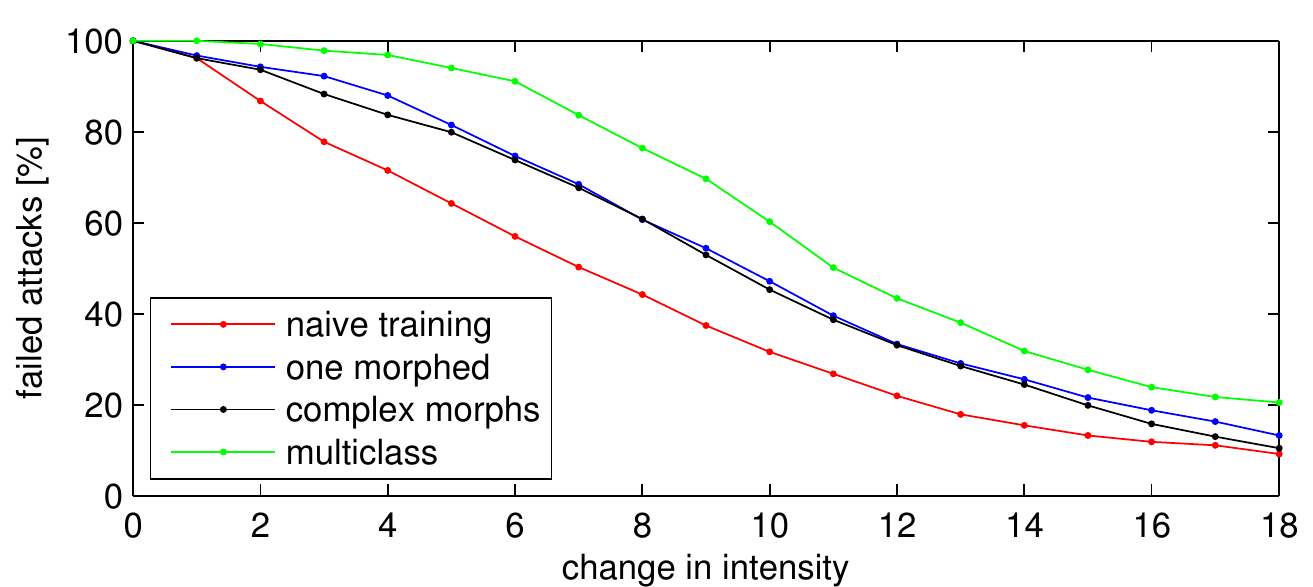}
\caption{Robustness against fast gradient sign attacks.}
\label{fig:DNNAttackStats}
\end{figure}

The other attack scenario we study is a black-box attack using adversarial examples. As mentioned above, we use the fast gradient sign method and a GoogLeNet for the substitute model to generate adversarial examples instead of the shallow architectures as in the original paper. Since an attacker would of course try to modify a morphing attack to let it be classified as an genuine image and not the other way around, we focus only on the classification results for morphing attacks. Figure \ref{fig:DNNAttackStats} shows the robustness against this kind of attack. All networks were vulnerable against this attack with a strong modification in the image, but especially for small and for the human eye not visible changes their vulnerability differ quite strong. The network that was trained using the naive method is the most vulnerable, the 2-classes training methods with partial morphs are more robust, but still perform worse than the multiclass training method. Even with a change of the intensities by $\pm6$, which leads to a visible noise-like pattern (see figure \ref{fig:DNNAttacksNaive}), still about 91\% of the adversarial examples are correctly classified. 

\subsubsection{Relevance Distribution}
\begin{figure}[!htpb]
\centering
\includegraphics[width=0.9\textwidth]{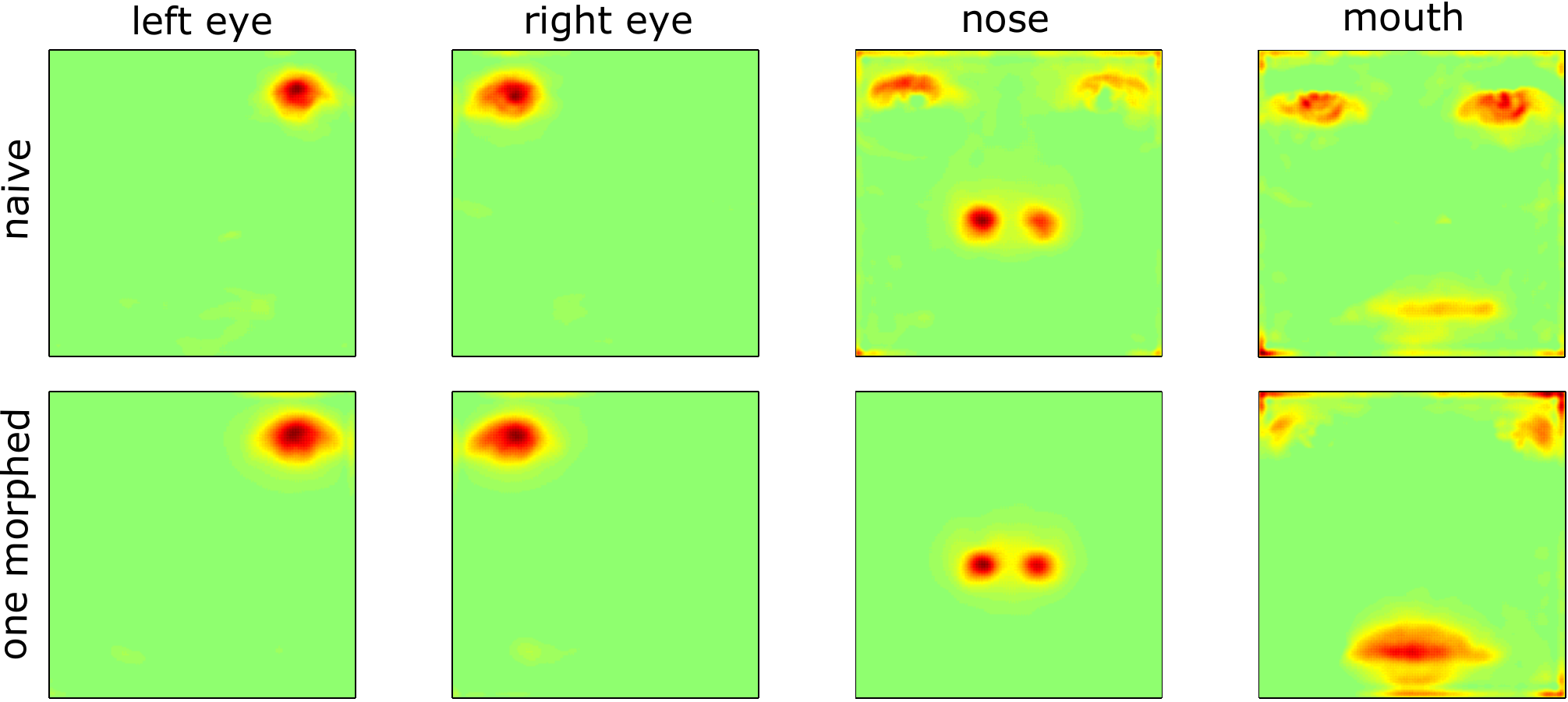}
\caption{Average mean adjusted positive relevance distribution for the naive and one morphed training method grouped by the different kinds of partial morphs.}
\label{fig:HeatmapsNaiveOneMorphed}
\end{figure}

Whereas the overall accuracy of our networks differs only less than 5 percent, we got quite different results for the robustness against partial morphs. The robustness also differs strongly for different kinds of partial morphs for the same network. Using relevance propagation, we try to get some insights about this behavior. We applied the relevance propagation on partial morphs and only if the softmax output of the network for the class morphing attack was at least 0.1, since if there is no or only little relevance in the neuron for the class "morphing attack", there will be no or only noisy relevance for this class in the image. Some neurons in the network are always active and thus lead to relevance that is always present. To overcome this problem, we subtracted the average relevance for each pixel. Finally, we removed all negative relevance, since negative relevance can be interpreted as inhibiting the decision for this class.\\
The relevance for the three networks that were trained directly on the binary decision is mostly in the morphed region as one would expect (see table \ref{tab:partMorphsStats} and figure \ref{fig:HeatmapsNaiveOneMorphed}). For these three networks, a strong correlation between the location of the relevance and the detection of the partial morphs can be found. For instance, the naive trained network performs worse on partial morphs of the nose and mount and the relevance for partial morphs of the nose or mouth are to a large extend outside of the morphed region. The same behavior can be observed for the training method with complex morphs and partial morphs of the mouth. \\
\begin{table}[!htpb]
\center
\caption{Relevance distributions.}
\begin{tabular}{l|cccc|cccc}
& \multicolumn{8}{c}{relative amount of relevance per region} \\
morphed \ \ & \multicolumn{4}{c}{naive} & \multicolumn{4}{c}{one morphed} \\
region&\ left eye\ &\ right eye\ &\ nose\ &\ mouth\ &\ left eye\ &\ right eye\ &\ nose\ &\ mouth\\
\hline
left eye  & 0.84 & 0.00 & 0.02 & 0.14 & 0.96 & 0.00 & 0.01 & 0.04\\
right eye & 0.00 & 0.91 & 0.05 & 0.05 & 0.00 & 0.92 & 0.01 & 0.07\\
nose      & 0.21 & 0.28 & 0.47 & 0.04 & 0.00 & 0.01 & 0.97 & 0.02\\
mouth     & 0.34 & 0.27 & 0.04 & 0.35 & 0.17 & 0.12 & 0.04 & 0.68\\\\
& \multicolumn{8}{c}{} \\
& \multicolumn{4}{c}{complex morphs} & \multicolumn{4}{c}{multiclass} \\
&\ left eye\ &\ right eye\ &\ nose\ &\ mouth\ &\ left eye\ &\ right eye\ &\ nose\ &\ mouth\\
\hline
left eye  & 0.98 & 0.00 & 0.00 & 0.02 & 0.00 & 0.98 & 0.00 & 0.01\\
right eye & 0.00 & 0.92 & 0.00 & 0.08 & 0.98 & 0.00 & 0.02 & 0.00\\
nose      & 0.02 & 0.03 & 0.92 & 0.02 & 0.01 & 0.10 & 0.19 & 0.70\\
mouth     & 0.06 & 0.00 & 0.41 & 0.53 & 0.11 & 0.18 & 0.58 & 0.13
\end{tabular}
\label{tab:partMorphsStats}
\end{table}

The relevance distribution indicates that the model for forged eyes of the naive network is not as bad as it seems when looking at the robustness against partial morphs. The bad robustness might be due to the fact that the naive trained network was trained only on complete morphs and genuine images and thus has no knowledge about how to classify an image that contains characteristics of both classes. On the other hand, the relative bad robustness against partial morphs of the nose and mouth and the bad distribution of relevance for these morphs suggests that the naive trained network represents these parts only weakly. This is not surprising, since the eyes might be sufficient to distinguish between morphs and genuine images in most cases.
\subsubsection{Relevance Distribution Interpretation for Multiclass Training}
The relevance distribution for the multiclass training methods seems at the first glance completely misleading, but might point out to a network that compares different structures. A structure in the neural network that explains this kind of relevance distribution can, for instance, rely mainly on a model for genuine images and be connected as follows. Assuming we have two neurons that have a strong activation, if the eyes of the individual in the image looks like a male Caucasian in his late twenties. One neuron represents the left eye and the other one the right eye. If one of the eyes does not fit to the model, e.g.,  due to a morphing artifact, it has a low activation. If we subtract the activation of one neuron from the activation of the other one, we get either a large negative or positive activation. The large negative activation would be clamped by the rectified linear unit, while the position activation would remain and indicate a morph. As a result, we have a path of strong activation from this neuron, to the area in the input image around the genuine eye and thus relevance in its region. Thus, the absence of relevance in a region can also indicate the importance of this region. For comparisons with feature maps that show a strong activation if a region is morphed, we can still expect that the relevance for a genuine region is different than if this region is forged.
\begin{figure}[!htpb]
\centering
\includegraphics[width=0.9\textwidth]{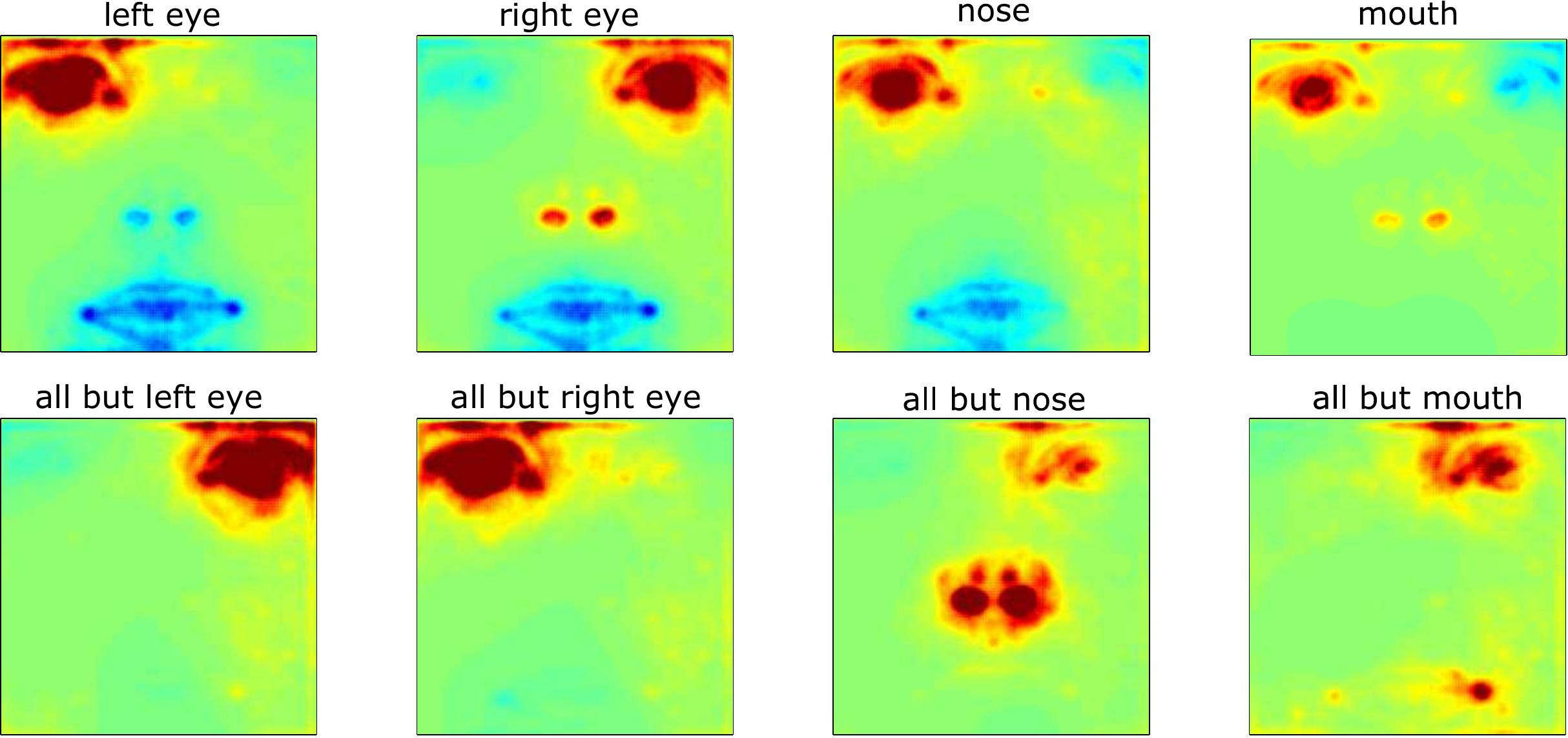}
\caption{Relevance distribution for the multiclass training of different kinds of partial morphs based on one input image pair.}
\label{fig:heatMapsPartMorphs}
\end{figure}

Figure \ref{fig:heatMapsPartMorphs} shows the relevance distribution of different partial morphs of the same pair of input images for the multiclass training method. In cases of one morphed region (upper images), there is nearly no relevance in the morphed region, but negative or positive relevance in the other parts of the image. If all but one region was morphed (lower images), there is again nearly no relevance in the forged parts, but a lot of relevance in the original part of the image. Some relevance is also assigned to the left eye if this region is forged, but it is much less than if this region is genuine.
\section{Conclusion}
In this paper, we presented different training methods based on modifications of the training data for the example of face morphing attack detection. We analyzed the resulting networks regarding accuracy, robustness and got insights about differences in their decision making.\\
We showed that accuracy does not always comes along with robustness against attacks on the network. Using partial morphs, in which only a part of the image is forged, during training, we enhanced the network to detect morphs also if only a part of the image was forged and we enhanced also the robustness against adversarial attacks. Our analysis on the decision making process of the networks indicates that the naive trained network focuses a lot on the region around the eyes and gives less priority to the region around the nose and mouth which leads to a weak distinctive model for these regions. The networks that were trained with partial morphs tend to have better models for all regions of the face, but have a worse accuracy. We overcome this problem by a complex multiclass pretraining by which we got a robust and accurate network. 
Our analysis of the decision making process of this network indicates that it compared different region among each other to detect morphing attacks, whereas this comparison structure was not visible for the other training methods.\\
In future work we will combine the complex multiclass pretraining with defence mechanisms against adversarial attacks \cite{SriArXiv18} to further robustify the neural network models.

\subsubsection*{Acknowledgements.} This work was supported by the German Federal Ministry of Education and Research (BMBF) through the Research Program ANANAS under Contract No. FKZ: 16KIS0511, and by the Fraunhofer Society through the MPI-FhG collaboration project ``Theory \& Practice for Reduced Learning Machines''.

\bibliographystyle{splncs}
\bibliography{egbib}
\end{document}